\documentclass[preprint, 12pt]{elsarticle}
\usepackage{graphicx}
\usepackage{subfig}
\usepackage{amssymb}
\usepackage{amsthm}
\usepackage{amsmath}
\usepackage{bm}
\usepackage[ruled, vlined]{algorithm2e}
\usepackage{url}
\journal{Computer Vision and Image Understanding}

\newtheorem{theorem}{Theorem}
\newtheorem{proposition}[theorem]{Proposition}
\newtheorem{definition}[theorem]{Definition}

\begin{document}

\begin{frontmatter}



\title{Selective Image Super-Resolution}

\author[a,b]{Ju Sun}
\author[b]{Qiang Chen}
\author[b]{Shuicheng Yan}
\author[b]{Loong-Fah Cheong}
\address[a]{Institute of Interactive \& Digital Media, National University of Singapore}
\address[b]{Department of Electrical and Computer Engineering, National University of Singapore}

\begin{abstract}
In this paper we propose a vision system that performs image
Super Resolution (SR) with selectivity. Conventional SR
techniques, either by multi-image fusion or example-based
construction, have failed to capitalize on the intrinsic
structural and semantic context in the image, and performed
``blind'' resolution recovery to the entire image area. By
comparison, we advocate example-based selective SR whereby
selectivity is exemplified in three aspects: region selectivity
(SR only at object regions), source selectivity (object SR with
trained object dictionaries), and refinement selectivity (object
boundaries refinement using matting). The proposed system takes
over-segmented low-resolution images as inputs, assimilates recent learning
techniques of sparse coding (SC) and grouped multi-task lasso (GMTL),
and leads eventually to a framework for joint figure-ground
separation and interest object SR. The efficiency of our framework
is manifested in our experiments with subsets of the VOC2009 and
MSRC datasets. We also demonstrate several interesting vision
applications that can build on our system.
\end{abstract}


\end{frontmatter}
\section{Introduction}
\label{sec:intro}
Super-resolution image reconstruction is the process to recover
a high-resolution image from a single or multiple low-resolution
input images~\cite{park2003spm}. In frequency domain, this corresponds to resolving
the beyond-Nyquist high-frequency components from the aliased
version of the spectrum ~\cite{tsai1984multiple}.
Apparently SR problem is under-determined by its nature, because practically
many high-resolution images can produce the same low resolution
image(s). Therefore it comes without surprise that the extensive
research on SR has worked on providing additional constraints
and/or incorporating various prior knowledge.
\begin{figure}[!tbp]
  \includegraphics[width=\linewidth]{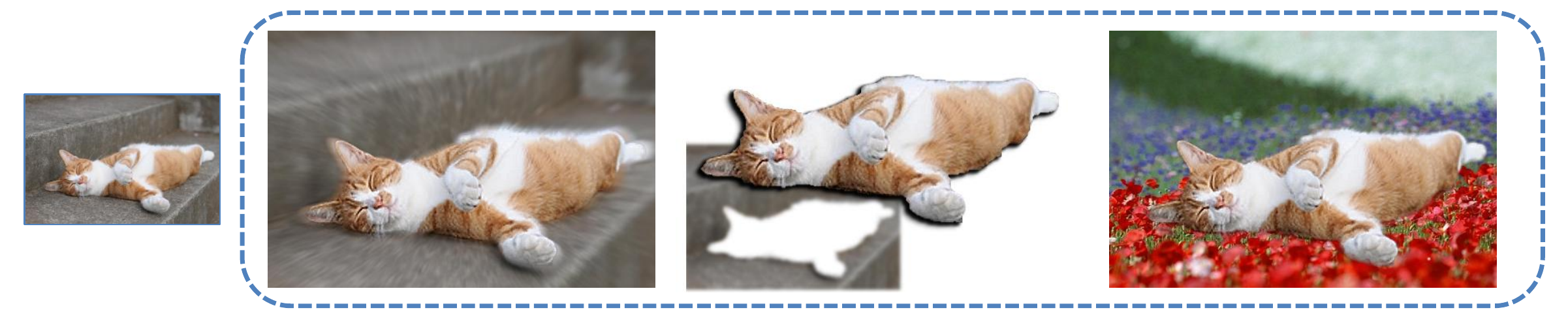}
  \caption{\textbf{Visual applications of our selective SR system}. (From left to right) zoom blurring, object pop-up, and image composition. (For better view, please refer to the electronic version)}
  \label{fig:top}
\end{figure}
Accordingly
existing SR techniques can be broadly classified into two
categories: 1) the classical multi-image fusion, and 2) the
sophisticated example-based construction.

Multi-image fusion techniques normally require multiple images of
the same scene with subpixel relative displacements as inputs. When sufficiently many such images are provided, direct SR reconstructions are feasible in both spatial and frequency
domains via solving systems of linear equations. Even
insufficient, these input images can be incorporated into
explicit regularization frameworks with various kind of prior
knowledge (mostly smoothness). A good review of all these
techniques is provided in~\cite{park2003spm}.
\begin{figure}[!tbp]
\begin{center}
\includegraphics[width=0.9\linewidth]{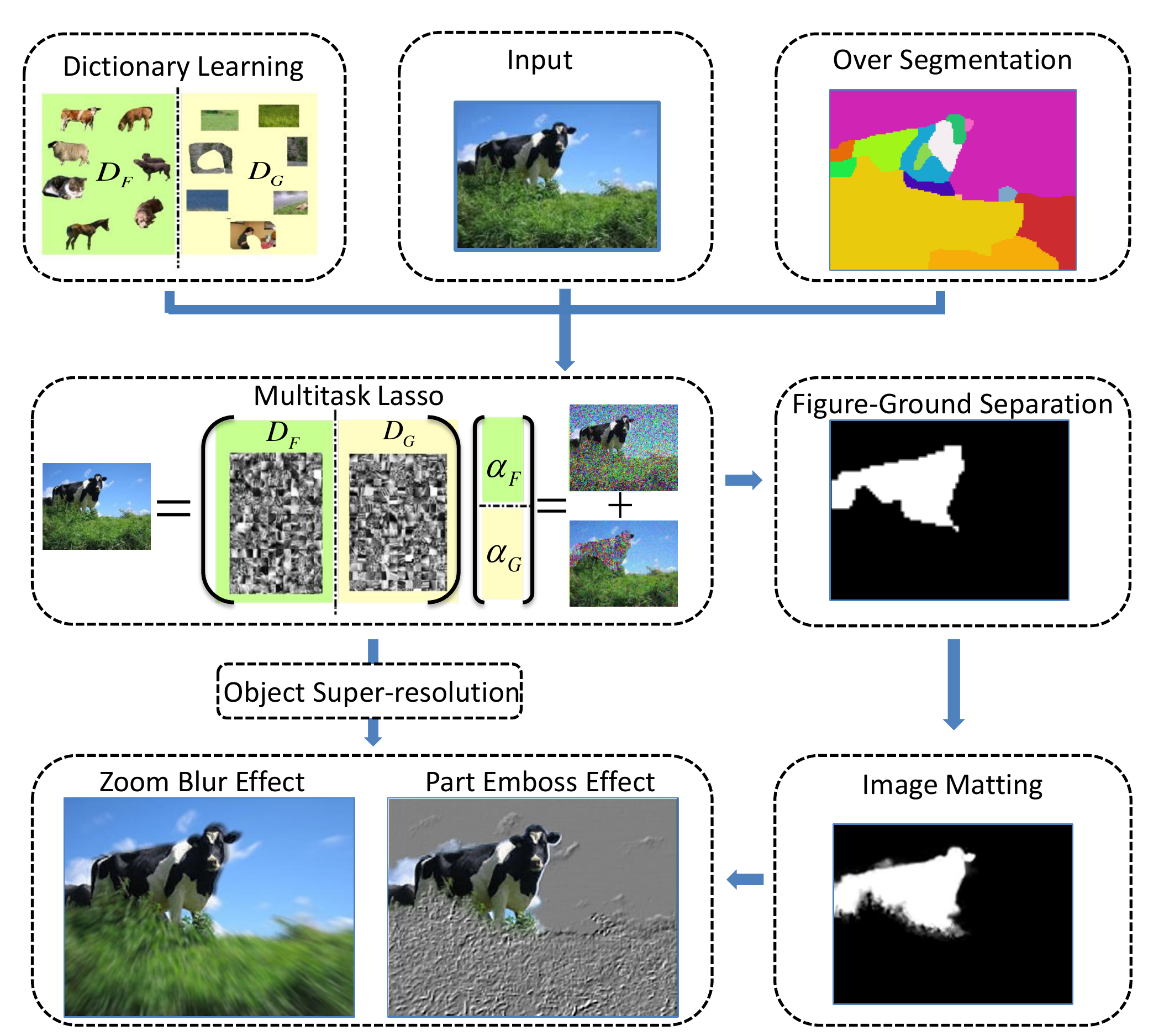}
\end{center}
  \caption{\textbf{Overview of our selective image SR system.} Given a low-resolution input image, our system first performs over-segmentation, and then employs GMTL (based on pre-trained dictionaries) to decide whether each segment belongs to the object region, leading to figure-ground separation. Image matting is used to refine the separation. It finally constructs super-solved version of the object region with the object dictionaries, and other special effects (zoom blur, part emboss) can be implemented by processing either the foreground or background region. }
\label{fig:overview}
\end{figure}
This family of SR algorithms are attractive due to the
simplicity of algorithmic implementation, and the ease of
multi-camera imaging and video capturing as inputs. As
demonstrated by many authors \cite{baker2002limits,
lin2004fundamental}, however, fusion-based SR can only provide numerically
less than double magnification factor. This has severely limited
their use for many applications.

This limitation has been broken since the introduction of example-base
SR techniques (or \textit{Image Hallucination}) \cite{baker2002limits, freeman2000learning}. Techniques in this vein feature
learning low- and high-resolution image patches from a
collection of low- and high-resolution image pairs.
Upon completion of the learning phase, early developed
algorithms (\textit{e.g.} \cite{baker2002limits,
freeman2000learning, freeman2002example,
baker2000hallucinating}) involve finding matched high-resolution patch for each low-resolution input and simply taking the high-resolution
as the recovered. These selection-based methods normally
entail large training datasets to ensure the desired
high-resolution patches can be best approximated by existing
training patches. By comparison, recent
developments (c.f. \cite{chang2004super,
ni2007image, yang2008image}) have introduced additional
flexibility and scalability by treating the patch generation
process as regression problems over properly trained/selected
basis patches. This innovation has made SR accessible to a wide range of applications with training sets of manageable sizes.

Despite the credible success SR research has made over the
decades, it is misfortune of SR being treated isolated as a
simple image enhancement process. In view of the integral nature of vision research, it is reasonable that one starts to link SR with other computer vision tasks, such as figure/ground separation, weak object identification. In fact,
it has been realized before that ``blind'' SR reconstruction to
the whole image area may not work well, in terms of
\textit{e.g.} resulting in over-smooth edges and corners
\cite{sun2003image}. To mitigate the adverse effect, SR
algorithms may be made adaptive to specific image regions. In
other words, the intrinsic image structure (such as edges,
region segments) and even the semantic context of image regions
(such as spatial layout, figure/ground separation, localized
labels for regions) need to be explicitly accounted for. This
coincides with the ongoing theme of recognition-oriented vision
research, namely building the synergy between shape and
appearance modeling (\textit{e.g.} epitomic shape and appearance
modeling \cite{jojic2003epitomic} and region-based recognition
\cite{gu2009recognition}). Furthermore, the coincidence sheds
light on integration of SR reconstruction with
other typical vision tasks, such as segmentation, recognition.

Moving towards this direction, we propose an image SR system
with selectivity. Compared to conventional SR algorithms,
selectivity is exercised in three modules of the system: 1)
perform SR only at object region(s) and blur out the background
region (region selectivity), 2) generate super-resolved patches
from training dictionaries of objects rather than background (dictionary selectivity), and 3) refine the
figure/ground boundaries with alpha matting techniques
\cite{wang2007image, levin2008closed} (refinement selectivity).
Three levels of selectivity has been facilitated by neat integration/adaptation of several state-of-the-art techniques, including image over-segmentation, sparse coding dictionary learning, multitask lasso, and image matting. Overall, the proposed system is able to separate the foreground objects out of the image region, perform SR to the object region and yield visually pleasing results after matting and other simple post-processings. Figure~\ref{fig:top} provides three special visual effects for a single input low-resolution image. Figure~\ref{fig:overview} gives an overview of our system pipeline and various techniques involved.

\section{Related Work}
The current work follows the line of example-based SR, which
dates back to the seminal papers \cite{freeman2000learning,
lin2004fundamental}. In these pioneering works, techniques such
as Markov random field (MRF) with belief propagation (BP) and
hierarchical nearest neighbor matching were used to establish
the low- to high-resolution patch correspondences. Thereafter,
\cite{sun2003image} introduced the primal sketch priors to
improve the reconstruction at the blurred edges, ridges, and
corners. All these early example-based SR techniques assume that
the recovered patches are to be selected from the training
datasets. This inevitably requires a large-scale datasets for
satisfactory patch approximation and hence reconstruction. This
limitation of the \emph{selection} paradigm triggers the
development of the \emph{regression} paradigm, in which novel
patches can be generated from a (usually linear) combination of
the existing. In this aspect, \cite{chang2004super} and
\cite{yang2008image} used respectively a linear combination of
several nearest neighboring patches and a sparsest combination
of training patches to approximate the input low-resolution
patch, and hence the low- and high-resolution correspondences.
Our work is inspired by \cite{yang2008image} which applied
sparsity priors to SR.

Sparse coding traces its root in signal processing and
compression\footnote{Hence it is also termed as
``compressive/compressed sensing'' in the signal processing
community. }, and is very recently introduced to the vision
community for face recognition \cite{wright2009robust} and many
other vision tasks \cite{sparsecvpr}. Recent research in
learning techniques has extended sparse coding to multiple-task
or multi-label decision scenarios \cite{zhang2006probabilistic,
yuan2006model, liu2009blockwise}, whereby sparsity is exploited
for feature selection and semantic inference. Another direction
of extension is on unsupervised or weakly supervised structure
discovery or over-complete coding dictionary learning
\cite{lee2007efficient, mairal2009online,
mairal2008discriminative}. These formulations explicitly require
learned visual structures or patterns (encoded as visual
dictionary parallel with the popular ``bag-of-features''
technique) facilitate sparse reconstruction. The current work
builds on both extensions and tunes them to our selective SR applications.

The theme of joint vision problem solving exhibited by the current work is not novel. In recognition-oriented vision research,
possibilities and advantages of joint visual object
segmentation, detection, and recognition and analysis have been suggested in many works, \textit{e.g.} the most recent ones
\cite{gu2009recognition, harzallah-combining}. Our proposal of
simultaneous object/background (or simply figure/ground)
separation and SR reconstruction
moves along the same line, hoping to discover new possibilities.
Our empirical results seem to provide us with positive answers.

Alpha matting is used for accurate
extraction of image foreground with transparent boundaries \cite{wang2007image,
levin2008closed}. One of the great challenges in matting is to
obtain an initial figure-ground separation map (the ``trimap'')
of decent accuracy. The output from figure-ground separation of
our algorithm (from the solution to the GMTL as
explained later) has fractional accuracy and can be used for
this purpose. Matting is used to fine-tune the boundary regions of our
highlighted foreground object.

\section{Image SR with Selectivity}
Our selective SR system stems from the confluence of computer
vision, machine learning and graphics techniques. Amongst them,
the hardcore mid-level vision research in image segmentation,
developments and applications of sparse coding and its
derivatives such as the multi-task Lasso and (sparsity-induced)
dictionary learning, and the extensively researched alpha
matting process, are integrated into our system. This part will
overview these techniques and provide some necessary details,
following the pipeline of our framework (as in Figure~\ref{fig:overview}).
\subsection{Image Segmentation}
Image segmentation has been one of the central topics in
mid-level computer vision, accounting for ``visual
grouping'' advocated by the Gestalt school of visual perception.
Literally, image segmentation is the partitioning of an image
into coherence groups of pixels, such that each group
corresponds to semantic-level objects or parts of objects. The
criterion for grouping is normally based on multiple image
cues, such as pixel intensity, color, texture, motion as
bottom-up driving force and categorical prior knowledge for
top-down modulation.

Albeit the extensive research on image segmentation ever since
the early vision days and the general belief that recognition
and alike high-level vision tasks should be founded on
segmentation, incorporating segmentation with other
high-level vision tasks has not seen much success. This is in part because
of the lack in a reliable and efficient segmentation algorithm
up to date, and fundamentally determined by the fact that visual
segmentation is inherently hierarchical and
task-driven. Whereas object-level segmentation in general relies also on the top-down
semantic-level knowledge, part-level segmentation tends to be
much easier as it is mainly driven by bottom-up low-level features. Hence the latter is much better researched and normally leads to ``over-segmentation''.

For our application, over-segmented image regions as the
unit blocks for figure-ground analysis carries over the structural and patch contextual information. We
assume homogeneity within each segmented region and the general
consistency in choosing source patches for SR reconstruction. For implementation, we
use the graph-based image segmentation of
\cite{felzenszwalb2004efficient}. This algorithm uses very
simple and intuitive measure for the local evidence of
boundaries, and makes greedy decisions of segmentation that
respect global structures and properties. Moreover, this
algorithm has approximately linear complexity
$\mathcal{O}\left(n\log n\right)$ \textit{w.r.t.} the number of edges, where $n$ is the number of image pixels (the graph is not fully connected).
\subsection{Sparse Coding and Dictionary
Learning} \label{dict_learn}
The input low-resolution image and its over-segments obtained above will work with dictionaries trained this part towards figure-ground separation via GMTL. The dictionary training is closely related to the sparse coding technique. In signal processing, sparse coding refers to the problem of
encoding signals (vector $\mathbf{y}$) as sparse (few nonzero
coefficients) as possible, over known or unknown basis (matrix
$\mathbf{D}\in \mathbb{R}^{p\times n}$, every column a
$p$-dimensional basis vector). The basis is usually overcomplete,
meaning $n\gg p$. If $\mathbf{D}$ is known, the problem is
normally formulated as locating the sparest coefficient vector
$\bm{\alpha}^*$ from the combinatorial
optimization\footnote{$\Vert\cdot\Vert_0$ is a pseudo-norm, and
simply counts the number of nonzero elements in a vector.}
\begin{equation}
\min_{\bm\alpha} \Vert\bm{\alpha} \Vert_0, \; \text{s.t.} \;
\mathbf{y}=\mathbf{D}\bm{\alpha}.
\end{equation}
The optimization is NP-hard, and in practice for $\bm\alpha$
with notable sparsity, it can be relaxed as a $\ell_1$ convex
optimization problem \cite{donoho2006most}
\begin{equation}
\min_{\bm\alpha} \Vert\bm{\alpha} \Vert_1, \; \text{s.t.} \;
\mathbf{y}=\mathbf{D}\bm{\alpha}.
\end{equation}
To account for practical noisy cases, the equality constraint is
normally relaxed as
$\Vert\mathbf{y}-\mathbf{D}\widetilde{\bm{\alpha}} \Vert_2^2\leq
\epsilon$, or in its Lagrangian form as
\begin{equation} \label{eq:lasso}
\min_{\widetilde{\bm\alpha}} \Vert\widetilde{\bm{\alpha}}
\Vert_1 +\lambda
\Vert\mathbf{y}-\mathbf{D}\widetilde{\bm{\alpha}}\Vert_2^2,
\end{equation}
which is commonly known to the statistical learning community as
Lasso and can be solved efficiently using the popular Least
Angle Regression (LAR, \cite{efron2004least}) method or the
stochastic cyclic coordinate descent method
\cite{friedman2007pathwise}.
%
Recently there is growing interest in the cases of sparse coding
without known basis. Intuitively, learning the basis from the
data can create more specialized basis system, and also reveals
the underlying structure and helps discover prototypical samples
from the data source alone (example applications in
\cite{lee2007efficient}). This is termed as ``(sparsity-induced)
dictionary learning''. Formally, to extract a dictionary
$\mathbf{D}\in \mathbb{R}^{p\times n}$ of size $n$ from $m$ data
samples $\left\{\mathbf{y}_i\right\}_{i=1}^m, \mathbf{y}_i\in
\mathbb{R}^p$ ($n<m$), the following optimization problem is to
tackle
\begin{equation}
\min_{\left\{\widetilde{\bm\alpha}_i\right\}_{i=1}^m,
\mathbf{D}} \; \; \sum_{i=1}^m
\left(\Vert\widetilde{\bm{\alpha}}_i \Vert_1 +\lambda
\Vert\mathbf{y}_i-\mathbf{D}\widetilde{\bm{\alpha}}_i\Vert_2^2\right).
\end{equation}
The objective is convex with respect to each group of the
variables, but not simultaneously. Hence one possible solution
scheme is to alternate between the dictionary learning,
\textit{i.e.} solving for $\mathbf{D}$ given
$\left\{\tilde{\bm\alpha}_i\right\}_{i=1}^m$, and the converse in the
consequent step. This is exactly the strategy suggested in
\cite{lee2007efficient}. In our application, we find the online
dictionary learning algorithm proposed in
\cite{mairal2009online} much more efficient. This online
solution is based on sequential solutions to quadratic local
approximation of the objective function and has proven convergence.
Our dictionary learning involves object images of $5$ known semantic classes and their pixel-level segmentation (detailed in experiment
part). Image patches (typically of size $3\times 3$ or $5\times
5$ pixels) are randomly sampled from both object regions and their
corresponding backgrounds. Sampled patches are used for training
a pair of sparsity-induced dictionaries of a particular object class and
the related backgrounds, respectively. For each dictionary,
high-resolution patches and their corresponding sub-sampled
low-resolution patches are trained together (by concatenating them and properly scaling each), to ensure the
consistency between the high- and low-resolution patch basis (as
was done in \cite{yang2008image}). These class-specific
dictionaries with figure-ground distinction essentially capture
the contextual relationship and occurrences of object and their
surroundings, and can help provide discrimination between object
and background regions, as what follows.

\begin{figure}[!tbp]
\begin{center}
\subfloat[Foreground
Dict]{\includegraphics[width=0.45\linewidth]{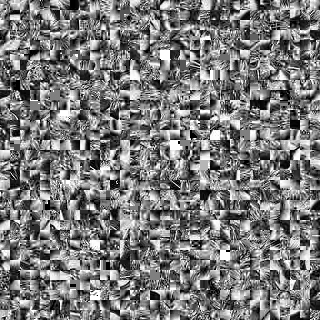}}\hspace{0.05in}
\subfloat[Background
Dict]{\includegraphics[width=0.45\linewidth]{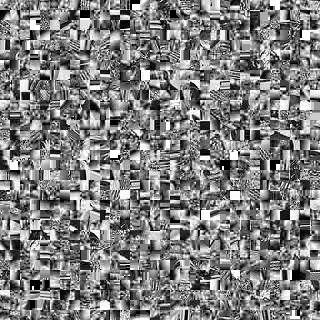}} \\
\subfloat[Dictionary Statistics]
{\includegraphics[width=0.98\linewidth]{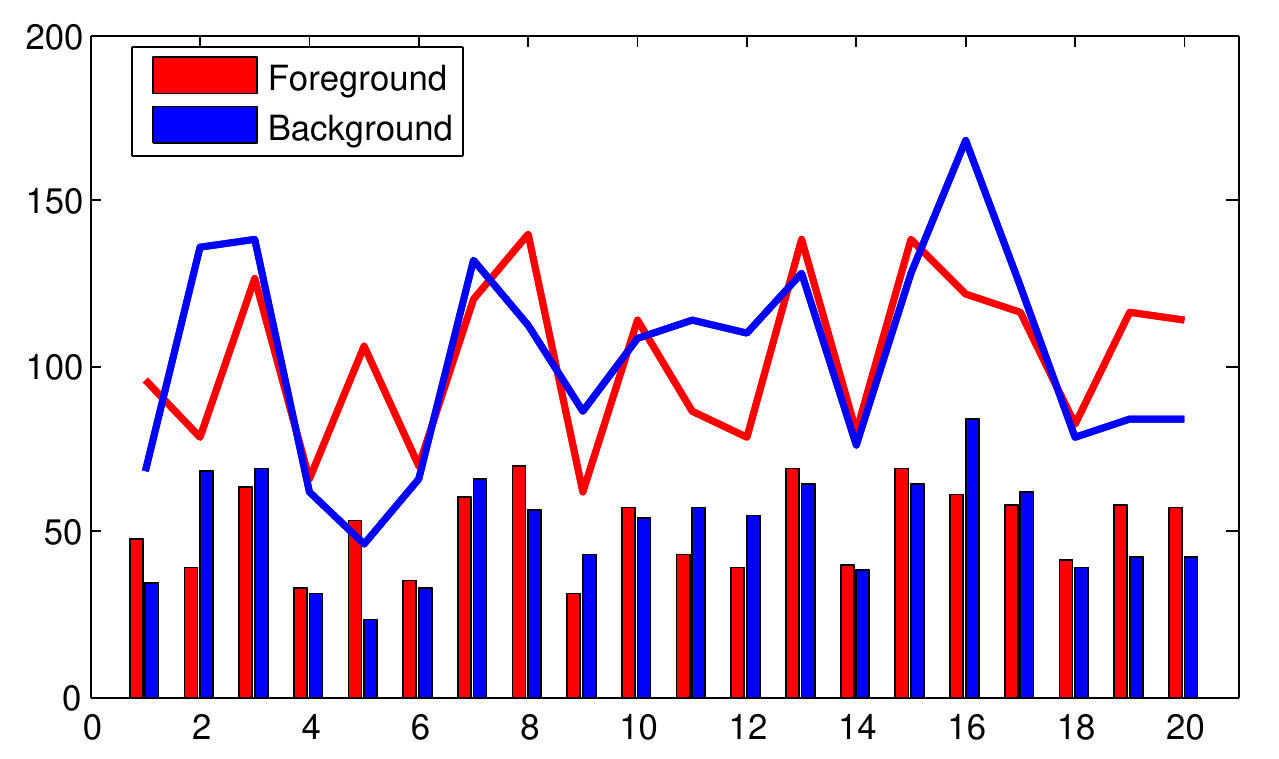}}
\end{center}
\caption{\textbf{Sparsity-induced dictionaries trained from
class-specific data.} The statistics are obtained by clustering these basis vectors (dictionary patches) into $20$ groups, and counting the fore- and back-ground patches within each group. The curves fit to these two sequences at doubled scale for clarity.}
\label{fig:dictionary}
\end{figure}

Figure~\ref{fig:dictionary} shows the foreground and background
dictionaries for a particular semantic class, and their distributions over $20$-bins. Note that these two dictionaries are not dramatically different. This is not surprising since the dictionary learning process tends to produce a generic basis, as noted by \textit{e.g.} \cite{lee2007efficient} and many others. Nevertheless, the local variations between the foreground and background distributions still exhibit notable difference.
\subsection{Grouped Multitask Lasso and Figure-Ground Separation}
Over-segmentation and dictionary learning essentially provide input for multitask Lasso in this part. Sparsity related techniques, such as Lasso in
Eq.~\eqref{eq:lasso}, are normally useful for identifying a
subset of most relevant features amongst a redundant collection
(shortly known as \emph{feature selection}). Recent developments
have extended Lasso into grouped setting (the Group Lasso
\cite{yuan2006model}, or GL), and multitask setting (the
MultiTask Lasso \cite{zhang2006probabilistic, liu2009blockwise},
or MTL). GL is designed to achieve group sparsity of variable
selection with respect to some pre-defined variable groups, and
is formulated as
\begin{equation} \label{eq:gl}
\min_{\bm\alpha} \Vert\mathbf{y}-\mathbf{D}\bm{\alpha}
\Vert_2^2+\lambda\sum_{g=1}^G \Vert
\bm{\alpha}_{\mathcal{I}_g}\Vert_2,
\end{equation}
where $\mathcal{I}_g$ is the index set of variable(s) belonging
to the $g^{th}$ group, for $g=1, \cdots, G$. On the other hand,
MTL aims to obtain the same sparsity pattern across tasks.
Assuming we are considering $K$ tasks together, and superscript
(such as $\mathbf{y}^{\left(k\right)}$) identifies a particular
task. We have the objective
\begin{equation}
\min_{\bm\Omega} \; \sum_{k=1}^K \Vert
\mathbf{y}^{\left(k\right)}-\sum_{j=1}^n
\omega_j^{\left(k\right)} \mathbf{d}_j^{\left(k\right)} \Vert_2^2
+ \lambda \sum_{j=1}^n \Vert \bm{\omega}_j\Vert_q,
\end{equation}
where $\mathbf{d}_j^{\left(k\right)}$ is the $j^{th}$ basis
vector for the $k^{th}$ task, and $n$ is the number of features
(size of the dictionary) as defined in last part. Moreover,
$\bm\omega_j= \left(\omega_j^{\left(1\right)}, \cdots,
\omega_j^{\left(K\right)}\right)^{\rm T}$ is the vector of all
coefficients for the $j^{th}$ feature across different tasks, and $\bm\Omega=\left(\bm\omega_1, \cdots, \bm\omega_n\right)^{\rm T}$. In
\cite{zhang2006probabilistic, liu2009blockwise}, the $q$-norm
for the coefficients is taken as the sup-norm, \textit{i.e.}
$\Vert \bm{\omega}_j \Vert_{\infty} = \max_k
\vert\omega_j^{\left(k\right)}\vert$. We argue that other valid
norms can also be taken, \textit{esp.} the $\ell_2$ norm. Taking
summation of $\ell_2$ norm as the penalization term not only
effectively helps combine MTL and GL as can be shortly seen, but
results in considerable computational saving as discussed below.

For our particular problem, we want to use the same dictionary
$\mathbf{D}=\left(\mathbf{d}_1, \cdots, \mathbf{d}_n\right)$ (concatenation of dictionaries of different object
classes, including both foreground and background ones) learned
from last part, and reconstruct local patches within each
segmented region simultaneously to identify if the region
belongs to the object or the background. Hence we drop the
superscript for the dictionary straight away and meanwhile
choose the $\ell_2$ norm. Then the objective we need to
optimize reduces to
\begin{equation}
\min_{\bm\Omega} \; \sum_{k=1}^K \Vert
\mathbf{y}^{\left(k\right)}-\sum_{j=1}^n
\omega_j^{\left(k\right)} \mathbf{d}_j \Vert_2^2 + \lambda
\sum_{j=1}^n \Vert \bm{\omega}_j\Vert_2.
\end{equation}
If we further enforce group sparsity as in GL, across tasks our
group reconstruction coefficients will take sub-matrices of the
matrix $\bm\Omega$. Hence the ultimate formulation will be
\begin{equation} \label{eq:mgl}
\min_{\bm\Omega} \; \sum_{k=1}^K \Vert
\mathbf{y}^{\left(k\right)}-\sum_{j=1}^n
\omega_j^{\left(k\right)} \mathbf{d}_j \Vert_2^2 + \lambda
\sum_{g=1}^G \Vert \bm{\Omega}_{\mathcal{I}_g}\Vert_F,
\end{equation}
where $\Vert\cdot \Vert_F$ is the Frobenius norm for matrices,
and $\mathcal{I}_g$ takes similar roles as in
Eq.~\eqref{eq:gl}.

Eq.~\eqref{eq:mgl} can be solved by (batch-mode) clockwise
coordinate descent as in \cite{liu2009blockwise}, with a
considerably large number of iterations. Instead, we turn the
regularization into a constraint and arrive at
\begin{equation}
\min_{\bm\Omega} \; \sum_{k=1}^K \Vert
\mathbf{y}^{\left(k\right)}-\sum_{j=1}^n
\omega_j^{\left(k\right)} \mathbf{d}_j \Vert_2^2 \; \text{s.t.}
\; \sum_{g=1}^G \Vert \bm{\Omega}_{\mathcal{I}_g}\Vert_F\leq C,
\end{equation}
where $C$ is a constraint parameter dual to the original
regularization parameter $\lambda$. By employing matrix
notations for $\mathbf{Y}=\left(\mathbf{y}^{\left(1\right)}, \cdots,
\mathbf{y}^{\left(K\right)}\right)$ in addition to $\mathbf{D}$ and $\bm\Omega$, the objective can be written as
\begin{equation} \label{eq:quad}
\min_{\bm\Omega} \; \bm{\Omega}^{\rm T}
\mathbf{D}^{\rm T}\mathbf{D}\bm\Omega-2\mathbf{Y}^{\rm T}\mathbf{D}\bm\Omega
\; \text{s.t.} \; \sum_{g=1}^G \Vert
\bm{\Omega}_{\mathcal{I}_g}\Vert_F\leq C.
\end{equation}
The following definitions and proposition will be important for
numerically solving the optimization.
\begin{definition}[Mixed $\ell_{p, q}$-Norm]
For a vector $\mathbf{x}\in \mathbb{R}^n$ and a set of disjoint
index set $\left\{\mathcal{I}_g\right\}_{g=1}^G$ such that
$\cup_g \mathcal{I}_g=\left\{1, \cdots, n\right\}$. The
$\ell_{p, q}$-norm for $\mathbf{x}$ is defined as $\Vert
\mathbf{x}\Vert_{p, q}=\left(\sum_g \Vert
\mathbf{x}_{\mathcal{I}_g} \Vert _q^p \right)^{1/p}$, where
$\mathbf{x}_{\mathcal{I}_g}$ is the tuple consisting of the
element{s} over indexes $\mathcal{I}_g$.
\end{definition}
\begin{definition}[$\ell_{p, q}$-Norm Balls]
$\mathcal{C}=\left\{\mathbf{x}\vert\Vert \mathbf{x}\Vert_{p,
q}\leq \tau\right\}$ is the $\ell_{p, q}$-norm ball of radius
$\tau$.
\end{definition}
\begin{proposition}[Projection onto an $\ell_{1, 2}$-Norm Ball of Radius $\tau$ \cite{schmidt2009optimizing}]
For a vector $\mathbf{x}\in \mathbb{R}^n$ and a set of disjoint
index set $\left\{\mathcal{I}_g\right\}_{g=1}^G$ such that
$\cup_g \mathcal{I}_g=\left\{1, \cdots, n\right\}$. The
Euclidean projection
$\mathcal{P}_\mathcal{C}\left(\mathbf{x}\right)$ onto the
$\ell_{1, 2}$-norm ball of radius $\tau$ is given by
\begin{equation}
\widetilde{\mathbf{x}_{\mathcal{I}_g}}=\mathop{\rm sgn}\left(\mathbf{x}_{\mathcal{I}_g}\right)\max\left(0, \Vert \mathbf{x}_{\mathcal{I}_g}\Vert_2-\lambda\right),
\end{equation}
where $\mathop{\rm sgn}\left(\mathbf{z}\right)$ is the signum function defined over the vector $\mathbf{z}$ as  $\mathop{\rm sgn}\left(\mathbf{z}\right)=\mathbf{z}/\Vert\mathbf{z} \Vert_2$, and $\lambda$ recursively defined over subset of the index set $\left\{\mathcal{I}_g\right\}_{g=1}^G$ as $\mathcal{I}_\lambda=\left\{j\in \left(1, \cdots, G\right)\vert \Vert \mathbf{x}_{\mathcal{I}_j} \Vert_2>\lambda\right\}$ and the constraint that $\sum_{j\in \mathcal{I}_\lambda} \left(\Vert \mathbf{x}_{\mathcal{I}_j}\Vert_2-\lambda\right)=\tau$.
\end{proposition}
$\lambda$ in the proposition can be solved efficiently \cite{duchi2008efficient}, and hence the projection. The $\ell_{1, 2}$-constrained quadratic optimization in Eq.~\eqref{eq:quad} has simple analytic gradients, and the constraint is a $\ell_{1, 2}$-norm ball with projection rule as discussed above. Hence we can employ the projected gradient method for convex optimization \cite{bertsekas-nonlinear} as described in Algorithm~\ref{algo:mtl}.
\begin{algorithm}[!htbp]
\caption{Group-Multitask Lasso Algorithm}
\label{algo:mtl}
\SetAlgoVlined  
Given $\mathbf{D}$, $\mathbf{Y}$, $C$, $\left\{\mathcal{I}_g\right\}_{g=1}^G$, $\eta$. Set $\bm\Omega^{\left(0\right)}$, $k\leftarrow 0$\\
\While {not Converged}{
\tcp{gradient descent}
$\bm\Omega^{\left(k+1\right)}=\bm\Omega^{\left(k\right)}-\eta\left(\mathbf{D}^T\mathbf{D}\bm\Omega^{\left(k\right)}-\mathbf{D}^T\mathbf{Y}\right)$\\
\tcp{vectorize submatrices}
\For{$g=1$ \emph{\KwTo} $G$}{
$\bm\beta_g$=$\text{vectorize}\left(\bm\Omega_{\mathcal{I}_g}^{\left(k+1\right)}\right)$
}
$\bm{\beta}=\left(\bm\beta_1^T;  \cdots;  \bm\beta_G^T\right)$ \\
\tcp{projection}
Solve $\lambda$ \\
\For{$g=1$ \emph{\KwTo} $G$}{
$\widetilde{\bm\beta_g}=\mathop{\rm sgn}\left(\bm\beta_g\right)\max\left(0, \Vert \bm\beta_g\Vert_2-\lambda\right)$ \;
$\bm\Omega_{\mathcal{I}_g}
^{\left(k+1\right)}=\text{devectorize}\left(\widetilde{\bm\beta_g}\right)$
}
}
\end{algorithm}

The convergence is typically within 50 iterations with the above projected
gradient method. Upon completion, we use the results to perform figure-ground separation. To this end, both the reconstruction error and the
reconstruction coefficients can be used. We find slightest
difference for them to distinguish between the object and
background, and hence we stick to the reconstruction
coefficients for simplicity. We directly compare the sum of
reconstruction coefficients within each semantic group
(foreground/background) to arrive at the figure-ground
separation. We note that we have scaled the dictionary elements
such that every feature vector has unity $\ell_2$ norm, and hence there
should not be cross-scale problem associated with the
reconstruction coefficients.
\begin{figure}[!tbp]
\begin{center}
\subfloat[Input
Image]{\includegraphics[width=0.45\linewidth]{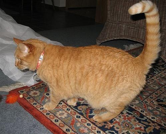}}
\hspace{0.05\linewidth}
\subfloat[Over-Segmentation]{\includegraphics[width=0.45\linewidth]{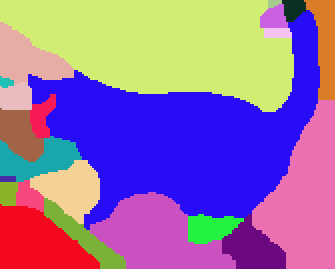}}
\\
\subfloat[FG Map-MTL]{\includegraphics[width=0.45\linewidth]{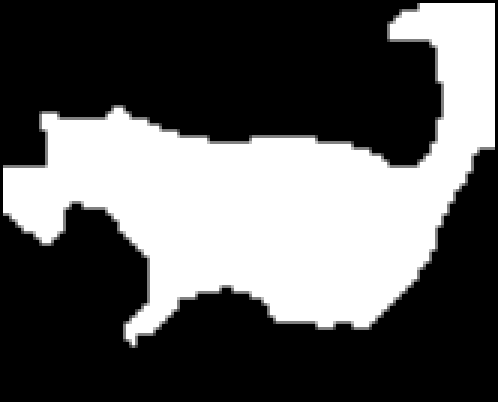}\label{FG_MTL}}
\hspace{0.05\linewidth}
\subfloat[FG Map-Lasso]{\includegraphics[width=0.45\linewidth]{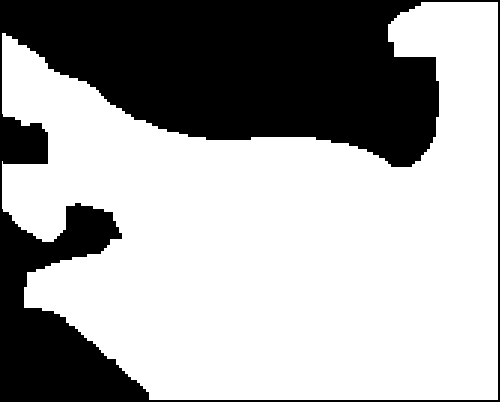}\label{FG_Lasso}}
\end{center}
\caption{\textbf{The figure-ground map produced by
segmentation-based GMTL vs. segmentation-based voting of patch-wise Lasso.} GMTL-based method produces better results. }
\label{fig:fg_map}
\end{figure}

Figure~\ref{fig:fg_map} presents one group of example results on figure-ground separation. Notice that the separation produced by this GMTL step (Figure~\ref{FG_MTL}) is not perfect, and is often dependent on the over-segmentation quality. Nevertheless, as compared to the results produced by patch-wise individual Lasso (Figure~\ref{FG_Lasso}), the former results are much better. This has illustrated the benefit of using GMTL for discrimination.

\subsection{Image Matting and Further Processing} \label{sec:matting}
Image matting as mentioned before is widely used for image
editing and other arts production applications. Mathematically,
matting involves the simultaneous estimation of the foreground
image $\mathbf{F}_z$ ($z$ denotes the pixel position) and the
background image $\mathbf{B}_z$ together with the alpha matte
$\bm\alpha_z$, given the observed image $\mathbf{I}_z$, and the
matting equation
\begin{equation}
\mathbf{I}_z=\alpha_z\mathbf{F}_z+\left(1-\alpha_z\right)\mathbf{B}_z.
\end{equation}
Matting is a typical inverse problem, and need additional
constraints or regularization to be solvable. Most existing
technique requires significant manual inputs which is undesirable
for our current work. Several recent algorithms need only sparse
user scribbles as inputs and even provide
closed-form solutions \cite{levin2008closed}.

We choose to use \cite{levin2008closed} to refine the figure-ground separation
map obtained from GMTL, and treat the map (fractional-values at
the boundaries) as the scribbled sparse alpha matte. The effectiveness of this novel employment of matting is confirmed by our empirical results (in experiment part). The solution of GMTL for each segment region can be used for SR reconstruction. In addition, we observe that following the patch-wise sparse reconstruction as discussed in \cite{yang2008image} provides additional performance gains. Furthermore, other image processing techniques can also be applied to the background region, leading to various visual applications (as shown in Figure~\ref{fig:top} and the upcoming Figure~\ref{fig:results}).
\section{Experiments and Discussions}
\subsection{Dataset Preparation}
We select images of five object categories: cow, horse, sheep, cat, and dog from the VOC2009 segmentation dataset \footnote{\url{http://pascallin.ecs.soton.ac.uk/challenges/VOC/voc2009/}} and the MSRC object class recognition\footnote{\url{http://research.microsoft.com/en-us/projects/objectclassrecognition/}} dataset (version 2), respectively. These datasets  are suitable for our purpose of object/background dictionaries training, because they provide pixel-level object/background segmentation groundtruth. We choose animal images to work with more diversity in textures. For each selected dataset, 15 images (about $10\%$ of the total) across all $5$ categories are used for testing, and the remaining for training.

Each training image and its down-sampled version (by the desired magnification factor, typically 3) constitute a high-/low-resolution image pair. $50, 000$ patches (with typical size $3\times 3$ pixels \textit{w.r.t.} the low-resolution image) for the object and the background respectively are then sampled from the training pairs, with the aid of the available groundtruth segmentation. For patch representation, we follow \cite{yang2008image} and use the first-order and second-order derivatives as features. The $1$-D filters used for feature extraction are
\begin{equation}
\mathbf{f}_1=\left[-1, 0, 1\right], \mathbf{f}_2=\mathbf{f}_1^{\rm T}, \mathbf{f}_3=\left[1, 0, -2, 0, 1\right], \mathbf{f}_4=\mathbf{f}_3^{\rm T}.  \notag
\end{equation}
Joint dictionary learning as described in Sec.~\ref{dict_learn} is then performed for each category class and its corresponding background, over the sampled patches. Each dictionary contains $1024$ basis patches.

\subsection{Figure-Ground Separation and Matting}
Based on the learned dictionaries, and the over-segmentation for an input image, the GMTL algorithm figures out the figure-ground separation based on the reconstruction coefficient vectors for each image segment. Several example output maps by this procedure is included in Figure~\ref{fig:results} (Group $C$ \footnote{Please refer to the legend in Figure~\ref{fig:results} and the caption therein for details about the groups.}). For comparison, we have also generated maps based on the voting of patch-wise reconstruction coefficients within each segment (single Lasso over each patch, and then voting within a segment, Figure~\ref{fig:results}, Group $B$). It is obvious that often GMTL produces more reliable object regions than the other way. This success is largely due to the joint solution to figure-ground separation within one segment, using the proposed GMTL technique. On the other hand, even GMTL often produces object map with fragmented boundaries or parts as evident from the examples. This is where matting comes into play. We apply our matting scheme as discussed in Sec-\ref{sec:matting} (results in Figure~\ref{fig:results}, Group $D$). Visual investigation suggests that matting does enhance the object boundaries much, notably at regions \textit{e.g.} the cow horns, the dogs' bodies and heads.
\subsection{SR and Other Visual Applications} \label{Appl}
\begin{figure*}[!tbp]
\begin{center}
\includegraphics[width=0.9\linewidth]{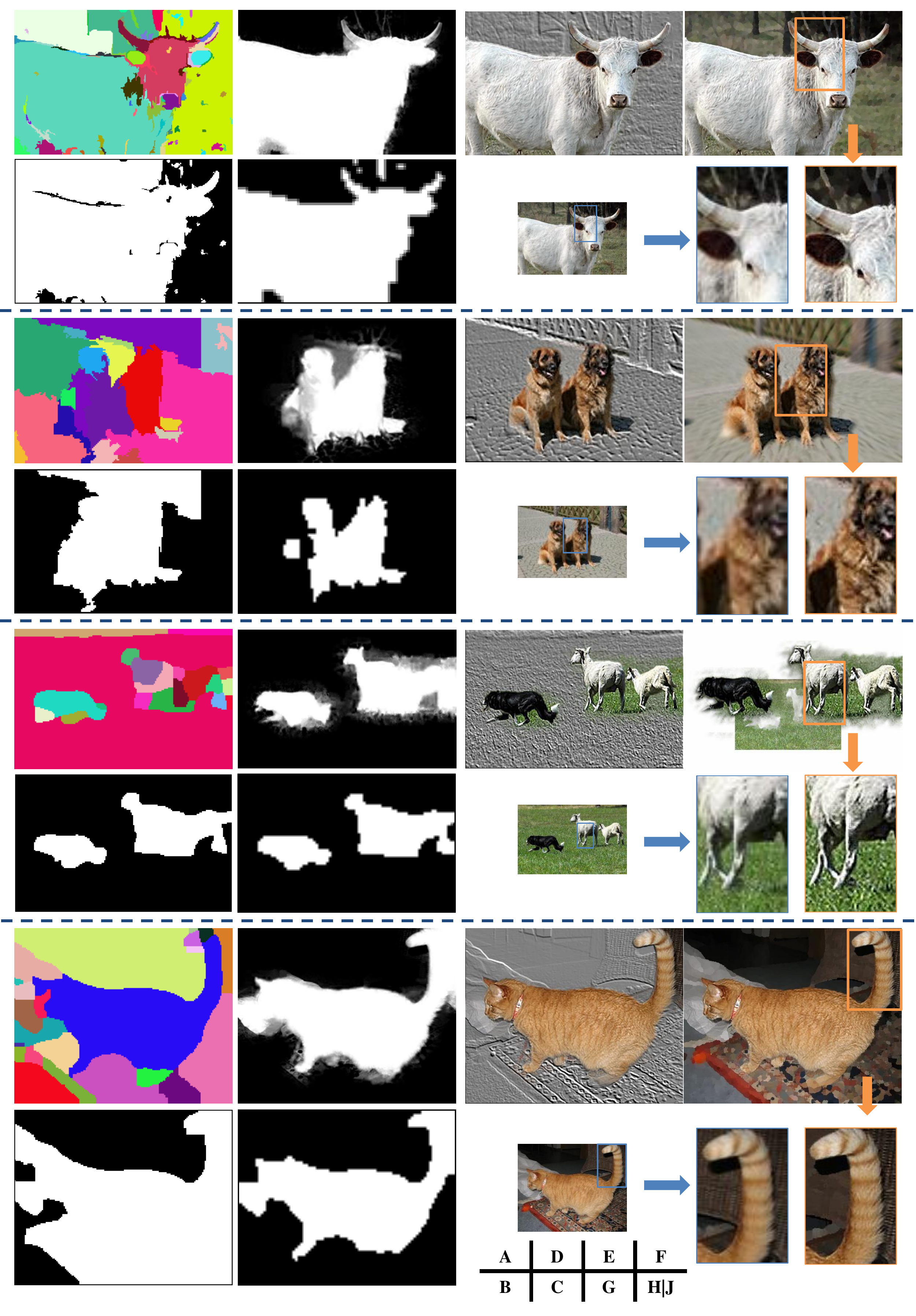}
\end{center}
   \caption{\textbf{Selective SR and various visual applications.} These are four comprehensive examples for demonstrating the system with the notations. $G$: input image, $A$: Over-segmentation map,  $B$: patch-wise figure-ground voting map within segments,  $C$: GMTL-based figure-ground map,  $D$: $C$ after matting, $E$ and $F$: special visual effects, $G$ and $F$: original patch and SR reconstructed patch. (For better view, please refer to the electronic version. Please zoom in to see the special effects.)}
\label{fig:results}
\end{figure*}
The SR reconstruction is hence based on the matted map instead of the original. Figure~\ref{fig:results}
compares several of the low-resolution textured patches/regions with that produced by our SR (Group $H$ vs. $J$). The reconstructed ones often contain significant more details than the original. Moreover, several possible visual effects by further processing the backgrounds or/and the objects are shown in Figure~\ref{fig:results} (Group $E$ and $F$).
\section{Conclusions and Future Work}
In this paper we employ and integrate several state-of-the-art methods in recent vision, learning, and graphics research, and build an SR system with selectivity that effectively jointly solves figure-ground separation and SR reconstruction. It is exciting to work along the classic over-segmentation algorithm with the sophisticated sparse coding and multitask Lasso techniques to achieve learning-based figure-ground separation. Equally exciting is the matting technique from graphics research that can effectively enhance the separation, and help us generate good SR reconstruction and fancy visual applications. We plan to further investigate the possibility of generic semantic class identification with the same setting.
\section*{Acknowledgment}
This work is partially supported by project grant NRF2007IDM-IDM002-069 on ¡°Life Spaces¡± from the IDM Project Office, Media Development Authority of Singapore, and partially by research grant NRF2008IDMIDM004-029 Singapore.
\section{Reference}
\bibliographystyle{SSR}
\bibliography{SSR}
\end{document}